\documentclass[letterpaper]{article} 
\usepackage{aaai2026}  
\usepackage{times}  
\usepackage{helvet}  
\usepackage{courier}  
\usepackage[hyphens]{url}  
\usepackage{graphicx} 
\usepackage{natbib}  
\usepackage{caption} 
\usepackage{algorithm}
\usepackage{algorithmic}

\usepackage{newfloat}
\usepackage{listings}

\usepackage{amsmath}                 
\usepackage{amssymb}
\usepackage{booktabs}
\usepackage{multirow}

\DeclareCaptionStyle{ruled}{labelfont=normalfont,labelsep=colon,strut=off} 
\floatstyle{ruled}
\newfloat{listing}{tb}{lst}{}
\floatname{listing}{Listing}
\nocopyright

\title{JanusNet: Hierarchical Slice-Block Shuffle and Displacement for Semi-Supervised 3D Multi-Organ Segmentation}
\author{
    Zheng Zhang\textsuperscript{\rm 1}, 
    Tianzhuzi Tan\textsuperscript{\rm 1}, 
    Guanchun Yin\textsuperscript{\rm 1}, 
    Bo Zhang\textsuperscript{\rm 2}, 
    Xiuzhuang Zhou\textsuperscript{\rm 1}
}
\affiliations{

    \textsuperscript{\rm 1} School of Intelligent Engineering and Automation, Beijing University of Posts and Telecommunications.\\
    \textsuperscript{\rm 2} State Key Laboratory of Networking and Switching Technology, School of Computer Science (National Pilot Software Engineering School), Beijing University of Posts and Telecommunications.\\
    zhangzheng@bupt.edu.cn, tan\_pros@bupt.edu.cn, yinguanchun@bupt.edu.cn, zbo@bupt.edu.cn, xiuzhuang.zhou@bupt.edu.cn
}

\usepackage{bibentry}

\DeclareRobustCommand{\legendsquare}[1]{%
  \textcolor{#1}{\rule{1.5ex}{1.5ex}}%
}
\definecolor{con1}{HTML}{000000}
\definecolor{con2}{HTML}{007F00}
\definecolor{con3}{HTML}{00FF00}
\definecolor{con4}{HTML}{FF0000}  
\definecolor{con5}{HTML}{0000FF}
\definecolor{con6}{HTML}{3399FF}
\definecolor{con7}{HTML}{FF69B4}
\definecolor{con8}{HTML}{7F00FF}

\definecolor{con9}{HTML}{FF7F00}
\definecolor{con10}{HTML}{A52A2A}
\definecolor{con11}{HTML}{A0522D}
\definecolor{con12}{HTML}{FF00FF}
\definecolor{con13}{HTML}{00FFFF}
\definecolor{con14}{HTML}{FFFF00}

\begin{document}

\maketitle

\begin{abstract}

Limited by the scarcity of training samples and annotations, weakly supervised medical image segmentation often employs data augmentation to increase data diversity, while randomly mixing volumetric blocks has demonstrated strong performance. However, this approach disrupts the inherent anatomical continuity of 3D medical images along orthogonal axes, leading to severe structural inconsistencies and insufficient training in challenging regions, such as small-sized organs, etc. 
To better comply with and utilize human anatomical information, we propose $\textbf{JanusNet}$, a data augmentation framework for 3D medical data that globally models anatomical continuity while locally focusing on hard-to-segment regions. Specifically, our $\textit{Slice-Block Shuffle}$ step performs aligned shuffling of same-index slice blocks across volumes along a random axis, while preserving the anatomical context on planes perpendicular to the perturbation axis. Concurrently, the $\textit{Confidence-Guided Displacement}$ step uses prediction reliability to replace blocks within each slice, amplifying signals from difficult areas. This dual-stage, axis-aligned framework is plug-and-play, requiring minimal code changes for most teacher-student schemes. Extensive experiments on the Synapse and AMOS datasets demonstrate that JanusNet significantly surpasses state-of-the-art methods, achieving, for instance, a 4\% DSC gain on the Synapse dataset with only 20\% labeled data.

\end{abstract}


\section{Introduction}

\begin{figure}[!tbp]  
  \centering
  \includegraphics[width=1\columnwidth]{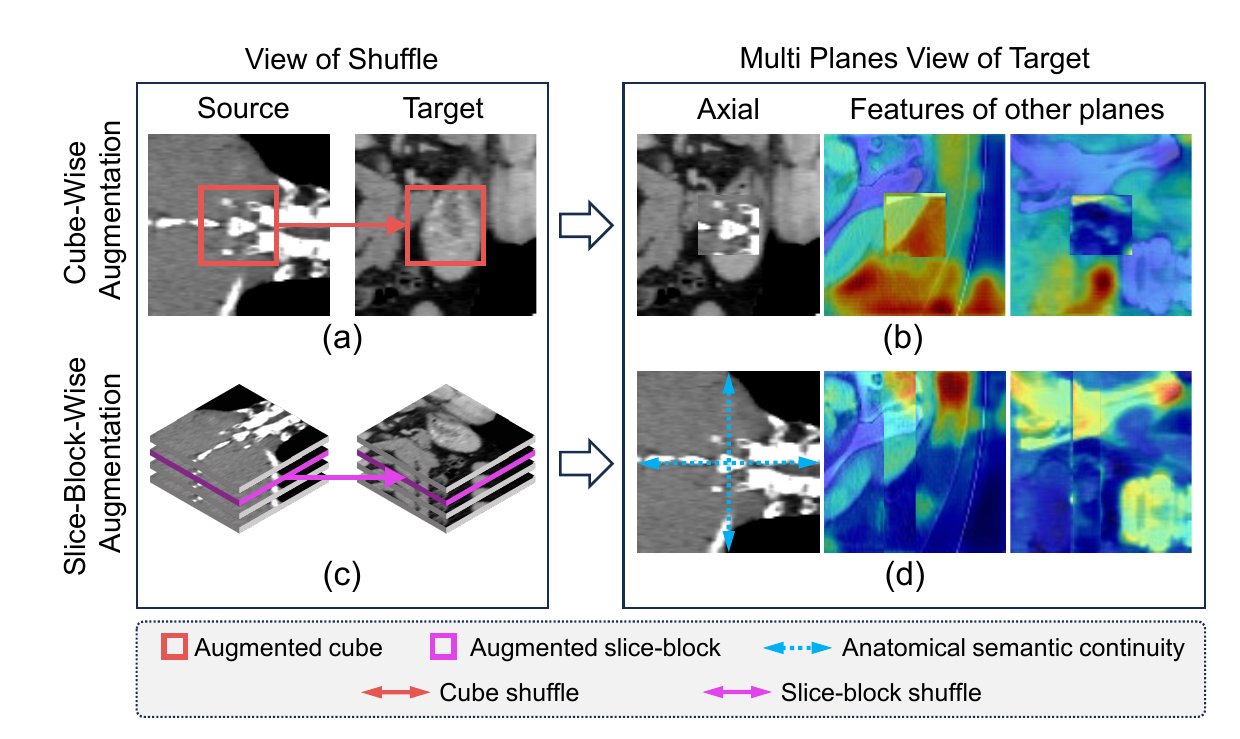}
  \vspace{-6mm}
  \caption{Illustration of augmentation at different levels. (a) Cube-wise single-step shuffle operation. (b) Cube-wise augmentation disrupts anatomical semantic continuity along all three axes. (c) Slice-block-wise single-step shuffle operation. (d) Slice-block-wise augmentation preserves anatomical structural consistency in planes orthogonal to the perturbed axis.}
  \vspace{-5mm}
  \label{fig:motivation}
\end{figure}



Despite the remarkable progress of fully supervised deep learning for medical image segmentation, its reliance on large-scale, high-quality annotations limits deployment\cite{tajbakhsh2020embracing,yang2023review}. Producing voxel-level labels requires domain expertise and is labor-intensive, whereas unlabeled data are abundant. Consequently, semi-supervised medical image segmentation has emerged as a compelling paradigm that reduces annotation cost and dependency while retaining strong performance potential.


Most semi-supervised approaches fall into two lines: self-training\cite{bai2017semi,yu2019ua-mt,bai2017semi} and consistency regularization\cite{du2023coarse,huang2022consistency3}. In self-training, a teacher trained on a small labeled subset generates pseudo-labels for unlabeled data, after which the network is optimized on the union. Consistency methods enforce prediction invariance between weakly and strongly augmented views. 
However, pseudo-labels derived from limited supervision are error-prone in the early stages, and naive use can induce \emph{confirmation bias}\cite{arazo2020pseudo} and lead the model to overfit incorrect targets. Moreover, labeled and unlabeled samples are often optimized in separate streams, leaving their objectives and losses misaligned and producing an \emph{empirical distribution mismatch}\cite{bai2023bcp}. Model-side remedies, such as temperature scaling, confidence thresholding, or loss re-weighting, mitigate but rarely remove this data-level mismatch, thereby constraining overall gains.


Multi-organ segmentation is more challenging than single-organ tasks. Large organs such as the liver and stomach occupy many voxels and exhibit stable textures, whereas small organs such as the adrenal glands, 
and elongated structures such as the esophagus, occupy few voxels and undergo stronger deformation, which leads to severe inter-class long-tailed distributions and pronounced scale disparity. In addition, many organs are adjacent and in contact, and their relative positions and topological relations are stable in three-dimensional space. Directly applying 2D or generic semi-supervised medical segmentation methods tends to introduce mismatches between semantics and position, which degrades performance. 
Thus, researchers try to inject anatomical priors on the model side \cite{kervadec2019boundary,shit2021cldice,hu2022boundary}. Although these techniques improve boundary quality and separability for certain structures, they still do not adequately address distribution mismatch between labeled and unlabeled data and the confirmation bias that arises between teacher and student models.


Medical volumetric imaging embodies stable anatomical priors\cite{cai2023orthogonal}, especially in 3D multi-organ settings where organ morphology evolves smoothly along the axial direction, relative layer positions are stable, and topological relations are well defined. To narrow the distribution gap between labeled and unlabeled data at the data level, position-aware mixing is a natural choice. Prior work \cite{bai2023bcp,chen2023magicnet} shows that relative-position-preserving cube mixing allows unlabeled samples to inherit organ semantics and layer information from labeled samples. However, many augmentation and perturbation methods are transferred from natural images, favoring tile-level reassembly or random copy-paste in 2D, and are then extended to cube-level perturbations in 3D volumes. Such practices are insufficient in three dimensions. As illustrated in Figure~\ref{fig:motivation}, cube-wise operations can disrupt anatomical continuity across axes.
In 2D tasks the negative effects of such discontinuity are relatively controllable due to limited pixel context and may even help diversity, but in 3D multi-organ segmentation, arbitrary reassembly of cubes that tends to break anatomical continuity, disrupt stable layer positions and topological relations among organs, and create mismatches between semantics and position. Small or elongated structures, such as the adrenal glands, gallbladder, esophagus, and vessels, are particularly vulnerable, with degraded recall and unstable boundaries, and pseudo-labels become less reliable in difficult regions.  Therefore, data augmentation should respect semantic continuity along the 3D axes and the priors on relative layer positions, so that the model can better learn complex organ morphology.


To reduce the impact of the above issues, we propose \textbf{JanusNet}, which applies aligned slice-block-level perturbations to 3D volumes. We partition a volume along a randomly chosen principal axis into consecutive slice blocks, then mix samples at the same layer index across volumes while preserving anatomical continuity on the planes orthogonal to that axis (see Figure~\ref{fig:motivation}). This narrows the distribution gap between labeled and unlabeled data at the data level and retains semantic continuity, providing useful priors for small organs and hard regions. JanusNet adopts a teacher-student framework and introduces two stage-wise, layer-aware augmentations built on the slice-block shuffle. The first stage enforces global layerwise alignment, and the second stage performs local in-layer refinement. The two stages act progressively and cooperatively, striking a balance between global structure and difficult local details. 
Our main contributions are as follows:

\begin{itemize}
    \item We introduce a slice–block shuffle strategy that mixes $N$ aligned layers across labeled and unlabeled volumes on a random axis, encouraging unlabeled data to inherit relative–position semantics while preserving anatomical continuity on the orthogonal planes to that axis. 
    \item We propose a confidence–guided displacement that amplifies patch semantics by replacing unreliable regions with confident counterparts, correcting errors and improving the quality of consistency learning. 
    \item Our method is plug–and–play and collaborates with diverse backbones and semi–supervised paradigms. Extensive experiments on multiple datasets demonstrate consistent, state–of–the–art improvements over prior art.
\end{itemize}

\begin{figure*}[!tb]
  \centering
  \includegraphics[width=0.85\textwidth]{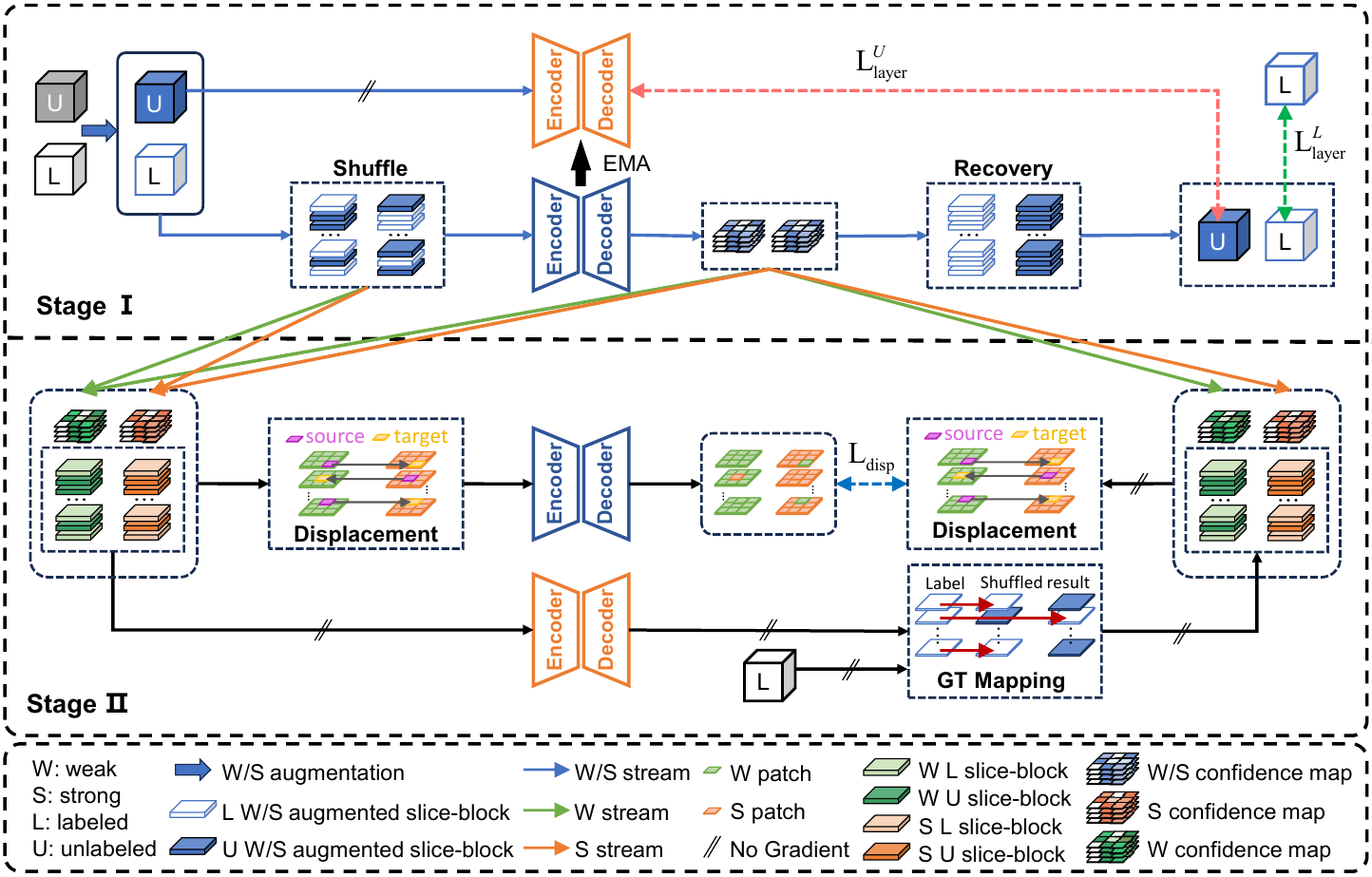}
  \vspace{-2mm}
  \caption{Overview of the proposed JanusNet framework, which 
  consists of two core steps, and adopts a teacher-student paradigm for pseudo-label supervision. 
  }
  \label{fig:framework}
  \vspace{-3mm}
\end{figure*}

\section{Related Work}

\subsubsection{Medical Image Segmentation.}

Accurate delineation of anatomical structures from CT/MRI underpins many computer–aided diagnosis and therapy pipelines. Existing methods broadly fall into two strands. The first focuses on 2D/3D architecture design\citep{ronneberger2015unet, milletari2016vnet, chen2021transunet, isensee2021nnu}. The second strand injects medical priors or weak supervision to enhance usability and generalization. Early work incorporated statistical shape templates, atlas registration, and topological constraints to regularize predictions toward anatomically plausible outputs\citep{wang2021preknow2, wang2020preknow3}. 
Despite these advances, many approaches still rely on extensive pixel-level annotations. This motivates semi-supervised formulations and data-level perturbations that better preserve anatomical semantics directions.

\subsubsection{Semi-supervised Medical Segmentation.}


Semi-supervised medical image segmentation (SSMIS) has chiefly evolved along two lines. 
\textbf{Consistency regularization} enforces prediction agreement on unlabeled volumes, typically under a teacher--student framework with weak perturbations. 
Representative methods include Mean Teacher \cite{tarvainen2017mean-teacher}  and its medical variants such as UA-MT \cite{yu2019ua-mt}, which leverage an exponential moving average (EMA) teacher together with uncertainty weighting to more effectively exploit unlabeled 3D data. 
Building on this idea, subsequent work explores richer perturbations and auxiliary tasks: Interpolation Consistency Training (ICT) \cite{verma2022interpolation} drives the decision boundary away from high-density regions via mixup-style interpolation. SASSNet augments UA-MT with signed distance map (SDM) regression to inject shape priors and improve boundary quality \citep{li2020shape}. DTC \citep{luo2021semi} imposes dual-task consistency (e.g., segmentation vs. boundary/distance cues) on unlabeled data to strengthen structural constraints.
\textbf{Pseudo-labeling and co-training} constitute the second line. 
Cross Pseudo Supervision (CPS) mitigates single-branch bias by exchanging pseudo labels between two learners and has become a popular baseline for segmentation \citep{chen2021cps}. 
Meanwhile, the FixMatch \citep{sohn2020fixmatch} paradigm-weak augmentation for high-confidence pseudo labels coupled with strong-augmentation consistency under confidence thresholding-has been adapted to semantic segmentation, inspiring a range of variants with confidence calibration and mutual-learning strategies.


\subsubsection{Data Perturbations for Semi-Supervising.}


Data perturbations are central to semi-supervised learning. In natural-image tasks, classic augmentations such as Cutout, MixUp, and CutMix regularize models \citep{cutout,zhang2017mixup,yun2019cutmix} by deliberately perturbing the inputs, thereby facilitating the use of unlabeled data. These perturbations have been embedded into semi-supervised frameworks: MixMatch \citep{berthelot2019mixmatch} produces low-entropy pseudo labels for unlabeled samples and mixes them with labeled data via MixUp \cite{zhang2017mixup}. ReMixMatch \citep{berthelot2019remixmatch} further introduces distribution alignment and augmentation anchoring. FixMatch \citep{sohn2020fixmatch} couples weak-augmentation pseudo labeling with strong-augmentation consistency under a confidence threshold, substantially narrowing the gap between supervised and unsupervised training .

Such strategies have been extended to medical image segmentation. BCP \citep{bai2023bcp} pastes labeled regions into unlabeled images and, symmetrically, unlabeled regions back into labeled images, reducing empirical distribution mismatch in both directions and yielding consistent gains across datasets. MagicNet partitions 3D volumes into $N^{3}$ cubes and performs ``partition--mix--recover,'' explicitly preserving relative-position priors so that consistency and small-organ recognition are strengthened across images and within volumes \citep{chen2023magicnet}. Beyond these, recent semi-supervised methods explore block or patch-level mixing or shuffling, such as Pair Shuffle Consistency \cite{he2024pair} and Double Copy-Paste \citep{bai2023bcp}, which further validate cross-image semantic mixing as an effective way to exploit unlabeled data. Yet perturbations at \emph{slice-block} level along an orthogonal axis remain underexplored.

\section{Method}

\subsection{Preliminaries}
\label{subsec:prelim}

Let the training set be $\mathcal{D}=\mathcal{D}_{L}\cup\mathcal{D}_{U}$ with $\mathcal{D}_{L}\cap\mathcal{D}_{U}=\varnothing$. 
Each labeled sample is a pair $(x,y)\in\mathbb{R}^{1\times D\times H\times W}\times\{0,\dots,C\}^{D\times H\times W}$ drawn from $\mathcal{D}_{L}$, and each unlabeled sample is $\bar x\in\mathbb{R}^{1\times D\times H\times W}$ drawn from $\mathcal{D}_{U}$. 
We adopt a teacher--student setting with student network $F(\cdot;\Theta_s)$ and exponential moving average (EMA) teacher $F_{\mathrm{ema}}(\cdot;\Theta_t)$. 
Given an unlabeled volume $\bar x$, we obtain a pseudo label $\tilde y=\arg\max\,\sigma(F_{\mathrm{ema}}(\bar x;\Theta_t))\in\{0,\dots,C\}^{D\times H\times W}$, where $\sigma(\cdot)$ denotes the softmax.

In each iteration, we form a mini-batch of size $B$ by sampling $B$ labeled and $B$ unlabeled volumes. 
A single orthogonal axis $a\in\{D,H,W\}$ is chosen at random and reused throughout the two steps in this iteration. 
Slice-Block Shuffle step produces the recovered predictions for labeled and unlabeled subsets, $(P_L^{\mathrm{rec}},P_U^{\mathrm{rec}})$, and yields the labeled and unlabeled \emph{layer} losses $\mathcal{L}^L_{\mathrm{layer}}$ and $\mathcal{L}^U_{\mathrm{layer}}$. 
Confidence-Guided Displacement step produces displaced inputs $\hat X^{\mathrm{disp}}$ with aligned displaced targets $\hat Y^{\mathrm{disp}}$, and yields the \emph{displacement} loss $\mathcal{L}_{\mathrm{disp}}$.

Our overall training objective combines these terms with scalar weights $\alpha,\beta\ge 0$:
{\small
\begin{equation}
\label{eq:prelim-total}
\mathcal{L}_{\mathrm{total}}
\;=\;
\mathcal{L}^L_{\mathrm{layer}}
\;+\;
\alpha\,\mathcal{L}^U_{\mathrm{layer}}
\;+\;
\beta\,\mathcal{L}_{\mathrm{disp}},
\end{equation}
}
where $\beta=\alpha\times \lambda_{\text{disp}}$, $\mathcal{L}^L_{\mathrm{layer}}=\ell_{\mathrm{dice}}(P_L^{\mathrm{rec}},Y_L)$ and $\mathcal{L}^U_{\mathrm{layer}}=\ell_{\mathrm{dice}}(P_U^{\mathrm{rec}},\tilde Y_U)$ follow Sec.~\ref{subsec:slice_block}, with $Y_L$ the ground-truth labels for the labeled subset and $\tilde Y_U$ the EMA pseudo labels for the unlabeled subset. 
The displacement loss $\mathcal{L}_{\mathrm{disp}}$ follows Sec.~\ref{subsec:conf_disp} and is the mean of voxel-wise multi-class cross-entropy and soft multi-class Dice between the student prediction on $\hat X^{\mathrm{disp}}$ and $\hat Y^{\mathrm{disp}}$. 
The teacher parameters $\Theta_t$ are updated via EMA of $\Theta_s$.

\subsection{Slice-Block Shuffle}
\label{subsec:slice_block}

\paragraph{Partition.}
Given a mini-batch with labeled and unlabeled volumes
\(X_L=\{x_i\in\mathbb{R}^{1\times D\times H\times W}\}_{i=1}^{B}\) and
\(X_U=\{x_i\in\mathbb{R}^{1\times D\times H\times W}\}_{i=1}^{B}\), we form the merged set
\(X_B=X_L\cup X_U=\{x_i\}_{i=1}^{2B}\).
We randomly choose an orthogonal axis \(a\in\{D,H,W\}\).
Let the length along axis \(a\) be \(L_a\) and pick a slice-block thickness \(p\) so that \(L_a=pN\).
Each volume is uniformly partitioned along axis \(a\) into \(N\) \emph{slice-blocks} as
\(x_i=\mathrm{cat}_{a}(x_i^{[1]},\dots,x_i^{[N]})\), which we denote by \(O_{\mathrm{part}}^{(a)}(\cdot)\).

\paragraph{Shuffle.}
To shuffle only the selected axis while preserving the anatomical context on the remaining two axes, we draw, for each layer index \(j\in\{1,\dots,N\}\), a column-wise permutation over the batch indices, forming
\(R\in\{1,\dots,2B\}^{2B\times N}\) with \(R_{:,j}\) a permutation of \(\{1,\dots,2B\}\).
Applying \(R\) layer-wise yields the shuffled set \(\hat X_B=\{\hat x_i\}_{i=1}^{2B}\) with
\(\hat x_i=\mathrm{cat}_{a}(x_{R_{i,1}}^{[1]},\dots,x_{R_{i,N}}^{[N]})\);
we denote this cross-slice-block shuffling by \(O_{\mathrm{shuf}}^{(a)}(X_B;R)\).

\paragraph{Recovery.}
To map predictions on the shuffled inputs back to the original batch order, we compute the column-wise inverse permutation \(S=\mathrm{Inv}(R)\in\{1,\dots,2B\}^{2B\times N}\) satisfying \(R_{S_{k,j},j}=k\).
Let the student be \(F(\cdot;\Theta_s)\) and its mixed features on \(\hat X_B\) be \(\{E_i\}_{i=1}^{2B}\), with \(E_i^{[j]}\) the sub-feature of slice-block \(j\).
We \emph{unmix} features by concatenating inverse-mapped slice-blocks,
\(\widetilde E_k=\mathrm{cat}_{a}(E_{S_{k,1}}^{[1]},\dots,E_{S_{k,N}}^{[N]})\),
denoted by \(O_{\mathrm{rec}}^{(a)}(\{E_i\};S)\).

\paragraph{Pipeline.}
Passing the shuffled inputs through the student and softmax, and then applying recovery and splitting back to labeled/unlabeled parts, we obtain:
{\small
\begin{equation}
\begin{aligned}
(P_L^{\mathrm{rec}},\,P_U^{\mathrm{rec}})
&= O_{\mathrm{split}}( O_{\mathrm{rec}}^{(a)}( \hat P;\, S) ),\\
\hat P
&= \sigma(F(Z;\Theta_s)),\\
Z
&= O_{\mathrm{shuf}}^{(a)}( O_{\mathrm{part}}^{(a)}(X_{\mathcal B});\, R ),
\end{aligned}
\label{eq:overall-layer}
\end{equation}
}
where \(\sigma(\cdot)\) denotes softmax and \(O_{\mathrm{split}}(\cdot)\) selects the first B entries for \(P_L^{\mathrm{rec}}\) and the remaining B for \(P_U^{\mathrm{rec}}\).

\paragraph{Losses.}
Let \(Y_L\in\{0,1,\dots,C\}^{B\times D\times H\times W}\) be voxel-wise ground truth for labeled volumes, and \(\tilde Y_U\) pseudo labels for unlabeled ones (e.g., from an EMA teacher).
Using multi-class Dice loss \(\ell_{\mathrm{dice}}\), we define:
{\small
\begin{equation}
\begin{gathered}
\mathcal{L}^L_{\mathrm{layer}}(B;\Theta_s) = \ell_{\mathrm{dice}}(P_L^{\mathrm{rec}},\,Y_L),\\
\mathcal{L}^U_{\mathrm{layer}}(B;\Theta_s) = \ell_{\mathrm{dice}}(P_U^{\mathrm{rec}},\,\tilde Y_U).
\end{gathered}
\end{equation}
}

\subsection{Confidence-Guided Displacement}
\label{subsec:conf_disp}

Confidence-guided displacement operates within each aligned layer produced in Sec.~\ref{subsec:slice_block}, reusing the same axis \(a\), block thickness \(p\) so that \(L_a=pN\), and an in‑plane grid of size \(n\times n\).
We stack the weak and strong streams along an extra stream dimension and denote by \(V\in\mathbb{R}^{B\times 2\times 1\times D\times H\times W}\) the input volumes, by \(Y\in\{0,\dots,C\}^{B\times 2\times D\times H\times W}\) the voxelwise labels, by \(C\in[0,1]^{B\times 2\times D\times H\times W}\) the confidence maps, and by \(G\in\{0,1\}^{B\times 2\times D\times H\times W}\) the supervision indicators with \(G=1\) on voxels that carry ground truth.

\paragraph{Patching.}
Within each layer, we tile the in-plane slice into an \(n\times n\) grid of patches, producing patchified tensors that preserve the layer index \(\ell\in\{1,\dots,N\}\) and grid coordinates \((u,v)\).
We denote this operation by \(O_{\mathrm{patch}}^{(a)}\), yielding
\(Z_{\mathrm{patch}}=O_{\mathrm{patch}}^{(a)}(V,Y,C,G)\) with shape \(\mathbb{R}^{B\times 2\times N\times n\times n\times(\cdot)}\).
\((\cdot)\) denotes the length of the last dimension, which depends on the tensor under consideration.

\paragraph{Statistics.}
For each patch identified by \((b,\ell,u,v)\) on each stream, we compute the mean confidence from \(C\) and derive a supervision flag from \(G\) indicating whether the patch contains any labeled voxels.
Comparing the stream-wise mean confidences gives high/low confidence indicators, and combining these with the supervision flag produces source and target candidates per stream and per location.
We aggregate these decisions into \(M_{\mathrm{stat}}=O_{\mathrm{stat}}(Z_{\mathrm{patch}})\), represented as two boolean masks \(M_{\mathrm{src}}\) and \(M_{\mathrm{tgt}}\) of shape \(\{0,1\}^{B\times 2\times N\times n\times n}\) that indicate, respectively, the selected source and target patches.

\paragraph{Top-\(K\) selection.}
For each sample \(b\) and each layer \(\ell\), we compute the inter-stream confidence gap on the \(n\times n\) grid and retain the \(K\) locations with the largest gaps to concentrate displacement on the most discriminative positions.
This yields a selection mask \(M_{K}=O_{\mathrm{topk}}(M_{\mathrm{stat}};K)\) with shape \(\{0,1\}^{B\times 1\times N\times n\times n}\), which is broadcast along the stream dimension during the subsequent swapping.

\begin{table*}[!t]
  \centering
  \footnotesize
  \setlength{\tabcolsep}{1.3mm}
   \begin{tabular}{@{}cr|cc|cccccccccccccc@{}}
    \toprule
    & \multirow{2}{*}{Method} & Avg. & Avg. & \multicolumn{13}{c}{Dice of Each Class}  \\
    & & Dice & ASD & Sp & RK & LK & Ga & Es & Li & St & Ao & IVC & PSV & Pa & RAG & LAG\\
    \midrule
    \multicolumn{1}{c|}{} & VNet (fully) 2016 & 62.09 ± 1.2 & 10.28 ± 3.9 & 84.6 & 77.2 & 73.8 & 73.3 & 38.2 & 94.6 & 68.4 & 72.1 & 71.2 & 58.2 & 48.5 & 17.9 & 29.0 \\
    \midrule
    \multicolumn{1}{c|}{\multirow{6}{*}{\rotatebox[origin=c]{90}{General}}}  & UA-MT 2019  & 20.26 ± 2.2 & 71.67 ± 7.4 & 48.2 & 31.7 & 22.2 & 0.0 & 0.0 & 81.2 & 29.1 & 23.3 & 27.5 & 0.0 & 0.0 & 0.0 & 0.0 \\
    \multicolumn{1}{c|}{} & URPC 2021  & 25.68 ± 5.1 & 72.74 ± 15.5 & 66.7 & 38.2 & 56.8 & 0.0 & 0.0 & 85.3 & 33.9 & 33.1 & 14.8 & 0.0 & 5.1 & 0.0 & 0.0 \\
    \multicolumn{1}{c|}{} & CPS 2021  & 33.55 ± 3.7 & 41.21 ± 9.1 & 62.8 & 55.2 & 45.4 & 35.9 & 0.0 & 91.1 & 31.3 & 41.9 & 49.2 & 8.8 & 14.5 & 0.0 & 0.0 \\
    \multicolumn{1}{c|}{} & SS-Net 2022  & 35.08 ± 2.8 & 50.81 ± 6.5 & 62.7 & 67.9 & 60.9 & 34.3 & 0.0 & 89.9 & 20.9 & 61.7 & 44.8 & 0.0 & 8.7 & 4.2 & 0.0 \\
    \multicolumn{1}{c|}{} & DST 2022  & 34.47 ± 1.6 & 37.69 ± 2.9 & 57.7 & 57.2 & 46.4 & 43.7 & 0.0 & 89.0 & 33.9 & 43.3 & 46.9 & 9.0 & 21.0 & 0.0 & 0.0 \\
    \multicolumn{1}{c|}{} & DePL 2022  & 36.27 ± 0.9 & 36.02 ± 0.8 & 62.8 & 61.0 & 48.2 & 54.8 & 0.0 & 90.2 & 36.0 & 42.5 & 48.2 & 10.7 & 17.0 & 0.0 & 0.0 \\
    \multicolumn{1}{c|}{} & MagicNet 2023  & 60.57 ± 2.5 & 22.48 ± 6.3 & \underline{82.5} & \underline{91.0} & 89.5 & 11.2 & 0.0 & 89.4 & \underline{62.7} & 77.6  & 79.0 & 66.1 & 47.3 & 36.8 & 54.3 \\
    \midrule
    
    \multicolumn{1}{c|}{\multirow{8}{*}{\rotatebox[origin=c]{90}{Imbalance}}}  & Adsh 2022  & 35.29 ± 0.5 & 39.61 ± 4.6 & 55.1 & 59.6 & 45.8 & 52.2 & 0.0 & 89.4 & 32.8 & 47.6 & 53.0 & 8.9 & 14.4 & 0.0 & 0.0 \\
    \multicolumn{1}{c|}{} & CReST 2021  & 38.33 ± 3.4 & 22.85 ± 9.0 & 62.1 & 64.7 & 53.8 & 43.8 & 8.1 & 85.9 & 27.2 & 54.4 & 47.7 & 14.4 & 13.0 & 18.7 & 4.6 \\
    \multicolumn{1}{c|}{} & SimiS 2022 & 40.0 ± 0.6 & 32.98 ± 0.5 & 62.3 & 69.4 & 50.7 & 61.4 & 0.0 & 87.0 & 33.0 & 59.0 & 57.2 & 29.2 & 11.8 & 0.0 & 0.0 \\
    \multicolumn{1}{c|}{} & Basak et al. 2022  & 33.24 ± 0.6 & 43.78 ± 2.5 & 57.4 & 53.8 & 48.5 & 46.9 & 0.0 & 87.8 & 28.7 & 42.3 & 45.4  & 6.3 & 15.0 & 0.0 & 0.0 \\
    \multicolumn{1}{c|}{} & CLD 2022  & 41.07 ± 1.2 & 32.15 ± 3.3 & 62.0 & 66.0 & 59.3 & \underline{61.5} & 0.0 & 89.0 & 31.7 & 62.8 & 49.4 & 28.6 & 18.5 & 0.0 & 0.0 \\
    \multicolumn{1}{c|}{} & DHC 2023  & 48.61 ± 0.9 & 10.71 ± 2.6 & 62.8 & 69.5 & 59.2 & \textbf{66.0} & 13.2 & 85.2 & 36.9 & 67.9 & 61.5 & 37.0 & 30.9 & 31.4 & 10.6 \\
    
    \multicolumn{1}{c|}{} & GenericSSL 2023  &60.88 ± 0.7 & 2.52 ± 0.4 & 85.2 & 66.9 & 67.0 & 52.7 & \textbf{62.9} & 89.6 & 52.1 & \underline{83.0} & 74.9 & 41.8 & 43.4 & 44.8 & 27.2 \\
    \multicolumn{1}{c|}{} & SKCDF 2025 & 64.27 ± 1.36 & \textbf{1.45 ± 0.09} & 79.5 & 72.1 & 67.6 & 59.8 & \underline{60.7} & \textbf{93.3} & 61.7 & \textbf{85.4} & 78.5 & 41.8 & \underline{50.9} & 46.4 & 37.8 \\

    \multicolumn{1}{c|}{} & GA-MagicNet 2024  & \underline{68.43 ± 0.5} & \underline{3.11 ± 0.2} & 81.4 & \textbf{92.4} & \textbf{90.8} & 33.5 & 53.3 & 89.1 & 60.9 & 79.1 & \underline{82.1}  & \underline{66.7} & 48.7 & \underline{50.3} & \textbf{61.4} \\
    \multicolumn{1}{c|}{} & JanusNet (Ours)  & \textbf{72.67 ± 1.2} & 3.82 ± 0.5 & \textbf{87.9} & \underline{90.2} & \underline{90.1} & 40.7 & 55.0 & \textbf{93.3} & \textbf{75.0} & 79.2 & \textbf{83.3}  & \textbf{71.4} & \textbf{62.5} & \textbf{55.7} & \underline{60.5} \\
    \bottomrule
  \end{tabular}
  \vspace{-2mm}
  \caption{Quantitative comparison on \textbf{20\% labeled Synapse dataset}. Methods are classified as 'General' or 'Imbalance' depending on whether it is designed for data imbalance. Organ abbreviations: Sp (spleen), RK (right kidney), LK (left kidney), Ga (gallbladder), Es (esophagus), Li (liver), St (stomach), Ao (aorta), IVC (inferior vena cava), PSV (portal \& splenic veins), Pa (pancreas), RAG (right adrenal gland), LAG (left adrenal gland). Average Dice and ASD scores are reported in the format of mean ± standard deviation over three independent runs. The best two results are highlighted \textbf{boldfaced} and \underline{underlined}.}
  \label{with sota on synapse}
  \vspace{-2mm}
\end{table*}

\paragraph{Bidirectional displacement.}
At each retained location \((b,\ell,u,v)\), a patch is eligible as a \emph{source} if it is either (i) low-confidence yet contains any ground-truth voxels, or (ii) high-confidence but contains no ground-truth voxels; conversely, a patch is a \emph{target} if it is (i) high-confidence and \emph{true}, or (ii) low-confidence and \emph{pseudo}. Formally, these conditions are already encoded in the masks \(M_{\mathrm{src}}\) and \(M_{\mathrm{tgt}}\) from Sec.~\ref{subsec:conf_disp} (``Statistics'') and further restricted to the Top-\(K\) locations by \(M_{K}\) (``Top-\(K\)'').
We then form stream-wise, one-to-one pairings only where the two streams complement each other \emph{at the same spatial index};
in both cases, the location must also be selected by \(M_{K}\).
Only these paired positions are exchanged, and all other positions remain unchanged.
We perform the swap for \emph{both} image and label patches (the masks are broadcast to each cubic patch).
Finally, we invert the patching along axis \(a\) to restore the original layer layout and fold the stream axis into the batch, yielding displaced volumes and labels
\((\hat X^{\mathrm{disp}},\hat Y^{\mathrm{disp}})=O_{\mathrm{disp}}^{(a)}(Z_{\mathrm{patch}};M_{\mathrm{src}},M_{\mathrm{tgt}},M_{K})\),
with \(\hat X^{\mathrm{disp}}\in\mathbb{R}^{2B\times 1\times D\times H\times W}\) and
\(\hat Y^{\mathrm{disp}}\in\{0,\dots,C\}^{2B\times D\times H\times W}\).

\paragraph{Losses.}
We supervise the displaced student predictions using labels that are transported by the \emph{same} patching--selection--displacement pipeline.
Concretely, we first form a composite label tensor on the two streams by mixing ground truth and EMA pseudo labels,
\(Y^{\star}=G\odot Y+(1-G)\odot\tilde Y\),
where \(\tilde Y=\arg\max\,\sigma(F_{\mathrm{ema}}(V;\Theta_{t}))\).
Applying the identical \(O_{\mathrm{patch}}^{(a)}\), statistics, Top-\(K\), and \(O_{\mathrm{disp}}^{(a)}\) to \((V,Y^{\star},C,G)\) yields the displaced labels \(\hat Y^{\mathrm{disp}}\) that are exactly aligned with \(\hat X^{\mathrm{disp}}\).
Let \(P^{\mathrm{disp}}=\sigma(F(\hat X^{\mathrm{disp}};\Theta_{s}))\) be the student prediction on displaced inputs.
We then optimize the standard hybrid segmentation loss:
{\small
\begin{equation}
\label{eq:loss-disp-final}
\mathcal{L}_{\mathrm{disp}}
=
\tfrac{1}{2}\,\ell_{\mathrm{ce}}\!\big(P^{\mathrm{disp}},\,\hat Y^{\mathrm{disp}}\big)
+
\tfrac{1}{2}\,\ell_{\mathrm{dice}}\!\big(P^{\mathrm{disp}},\,\hat Y^{\mathrm{disp}}\big),
\end{equation}
}
where $\ell_{\mathrm{ce}}$ denotes the voxel-wise multi-class cross-entropy.

\section{Experiments}

\begin{table*}[!t]
  \centering
  \footnotesize
  \setlength{\tabcolsep}{1.3mm}
   \begin{tabular}{@{}cr|cc|cccccccccccccccc@{}}
    \toprule
    & \multirow{2}{*}{Method} & Avg. & Avg. & \multicolumn{13}{c}{Dice of Each Class}  \\
    & & Dice & ASD & Sp & RK & LK & Ga & Es & Li & St & Ao & IVC & Pa & RAG & LAG & Du & Bl & P/U \\
    \midrule
    \multicolumn{1}{c|}{} & VNet (fully) 2016 & 76.50 & 2.01 & 92.2 & 92.2 & 93.3 & 65.5 & 70.3 & 95.3 & 82.4 & 91.4 & 85.0 & 74.9 & 58.6 & 58.1 & 65.6 & 64.4 & 58.3 \\
    \midrule
    \multicolumn{1}{c|}{\multirow{6}{*}{\rotatebox[origin=c]{90}{General}}}  & UA-MT 2019  & 42.16  & 15.48  & 59.8 & 64.9 & 64.0 & 35.3 & 34.1 & 77.7 & 37.8 & 61.0 & 46.0 & 33.3 & 26.9 & 12.3 & 18.1 & 29.7 & 31.6 \\
    \multicolumn{1}{c|}{} & URPC 2021 & 44.93 & 27.44 & 67.0 & 64.2 & 67.2 & 36.1 & 0.0 & 83.1 & 45.5 & 67.4 & 54.4 & 46.7 & 0.0 & 29.4 & 35.2 & 44.5 & 33.2 \\
    \multicolumn{1}{c|}{} & CPS 2021 & 41.08 & 20.37 & 56.1 & 60.3 & 59.4 & 33.3 & 25.4 & 73.8 & 32.4 & 65.7 & 52.1 & 31.1 & 25.5 & 6.2 & 18.4 & 40.7 & 35.8 \\
    \multicolumn{1}{c|}{} & SS-Net 2022 & 33.88 & 54.72 & 65.4 & 68.3 & 69.9 & 37.8 & 0.0 & 75.1 & 33.2 & 68.0 & 56.6 & 33.5 & 0.0 & 0.0 & 0.0 & 0.2 & 0.2 & \\
    \multicolumn{1}{c|}{} & DST 2022 & 41.44 & 21.12 & 58.9 & 63.3 & 63.8 & 37.7 & 29.6 & 74.6 & 36.1 & 66.1 & 49.9 & 32.8 & 13.5 & 5.5 & 17.6 & 39.1 & 33.1 \\
    \multicolumn{1}{c|}{} & DePL2022 & 41.97 & 20.42 & 55.7 & 62.4 & 57.7 & 36.6 & 31.3 & 68.4 & 33.9 & 65.6 & 51.9 & 30.2 & 23.3 & 10.2 & 20.9 & 43.9 & 37.7 \\

    \multicolumn{1}{c|}{} & MagicNet 2023 & 54.08 
 & 29.03 & \underline{80.0} & \underline{84.5} & \underline{86.1} & \underline{47.9} & 0.0 & 85.1 & 50.7 & \underline{81.7} & 69.3 & 57.2 & \underline{46.0} & 0.0 & \underline{40.8} & \underline{62.9} & 19.2 \\
    \midrule
    \multicolumn{1}{c|}{\multirow{8}{*}{\rotatebox[origin=c]{90}{Imbalance}}}  & Adsh 2022 & 40.33 
 & 24.53 & 56.0 & 63.6 & 57.3 & 34.7 & 25.7 & 73.9 & 30.7 & 65.7 & 51.9 & 27.1 & 20.2 & 0.0 & 18.6 & 43.5 & 35.9 \\
    \multicolumn{1}{c|}{} & CReST 2021 & 46.55 & 14.62 & 66.5 & 64.2 & 65.4 & 36.0 & 32.2 & 77.8 & 43.6 & 68.5 & 52.9 & 40.3 & 24.7 & 19.5 & 26.5 & 43.9 & 36.4 \\
    \multicolumn{1}{c|}{} & SimiS 2022 & 47.27 & 11.51 & 77.4 & 72.5 & 68.7 & 32.1 & 14.7 & 86.6 & 46.3 & 74.6 & 54.2 & 41.6 & 24.4 & 17.9 & 21.9 & 47.9 & 28.2 \\
    \multicolumn{1}{c|}{} & Basak 2022 & 38.73 & 31.76 & 68.8 & 59.0 & 54.2 & 29.0 & 0.0 & 83.7 & 39.3 & 61.7 & 52.1 & 34.6 & 0.0 & 0.0 & 26.8 & 45.7 & 26.2 \\
    \multicolumn{1}{c|}{} & CLD 2022 & 46.10 & 15.86 & 67.2 & 68.5 & 71.4 & 41.0 & 21.0 & 76.1 & 42.4 & 69.8 & 52.1 & 37.9 & 24.7 & 23.4 & 22.7 & 38.1 & 35.2 \\
    \multicolumn{1}{c|}{} & DHC 2023  & 49.53 & 13.89 & 68.1 & 69.6 & 71.1 & 42.3 & 37.0 & 76.8 & 43.8 & 70.8 & 57.4 & 43.2 & 27.0 & 28.7 & 29.1 & 41.4 & 36.7 \\
    \multicolumn{1}{c|}{} & GenericSSL 2023  & 50.03 & 5.21 & 73.1 & 76.0 & 76.5 & 29.1 & 44.9 & 82.5 & 49.0 & 72.8 & 61.7 & 48.5 & 30.2 & 19.7 & 36.4 & 32.9 & 18.2 \\
    \multicolumn{1}{c|}{} & SKCDF 2025 & 53.81 & 5.97 & 77.1 & 77.9 & 71.2 & 34.1 & \underline{50.4} & \underline{88.6} & 51.6 & 80.9 & 58.9 & 48.8 & 33.0 & 30.2 & 32.2 & 45.9 & 26.4 \\
    
    \multicolumn{1}{c|}{} &  GA-MagicNet 2024 & \underline{63.51} 
 & \underline{4.58} & 78.9 & \textbf{85.5} & \textbf{87.2} & \textbf{50.0} & 49.1 & 86.9 & \underline{56.2} & 83.4 & \underline{70.3} & \underline{57.4} & \textbf{49.1} & \textbf{40.8} & 38.3 & \textbf{71.6} & \underline{47.9} \\
 \multicolumn{1}{c|}{} &  JanusNet (Ours) & \textbf{63.99} 
 & \textbf{4.45} & \textbf{83.0} & 83.7 & 84.9 & 43.5 & \textbf{59.6} & \textbf{89.1} & \textbf{63.6} & \textbf{83.7} & \textbf{73.8} & \textbf{59.9} & 41.6 & \underline{34.7} & \textbf{47.1} & 61.4 & \textbf{50.3} \\
    \bottomrule
  \end{tabular}
  \vspace{-2mm}
  \caption{Quantitative comparison on \textbf{5\% labeled AMOS dataset}. Organ abbreviations: Sp (spleen), RK (right kidney), LK (left kidney), Ga (gallbladder), Es (esophagus), Li (liver), St (stomach), Ao (aorta), IVC (inferior vena cava), Pa (pancreas), RAG (right adrenal gland), LAG (left adrenal gland), Du (duodenum), Bl (bladder), P/U (prostate/uterus).}
  \label{with sota on amos}
  \vspace{-2mm}
\end{table*}

\begin{figure*}[!tbp]
 \centering
 \includegraphics[width=1.8\columnwidth]{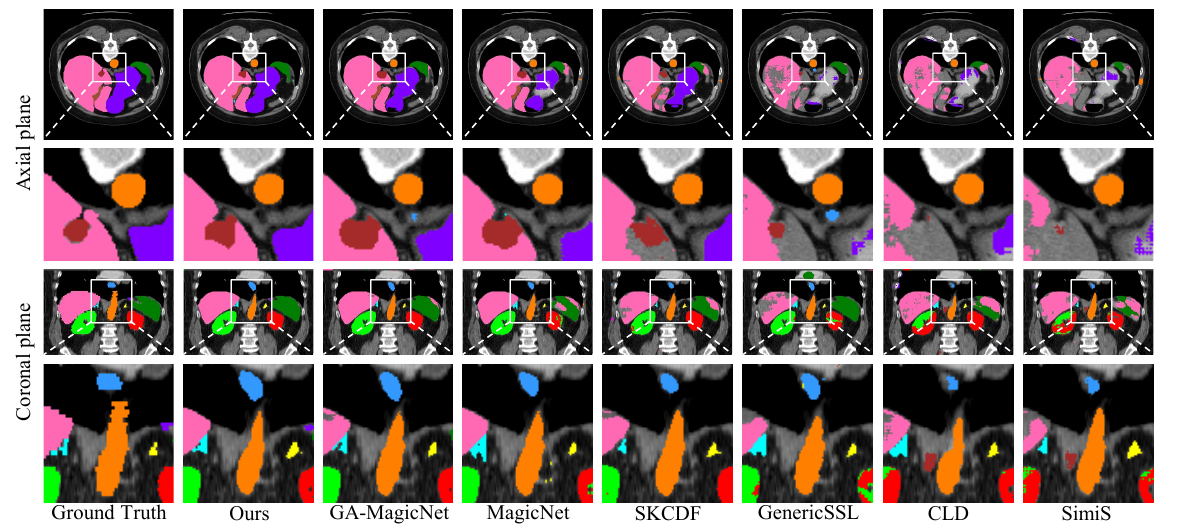}
 \vspace{-3mm}
 \caption{Visual comparison on the 20\% labeled Synapse dataset: \legendsquare{con2} spleen, \legendsquare{con3} right kidney, \legendsquare{con4} left kidney, \legendsquare{con5} gallbladder, \legendsquare{con6} esophagus, \legendsquare{con7} liver, \legendsquare{con8} stomach, \legendsquare{con9} aorta, \legendsquare{con10} inferior vena cava, \legendsquare{con11} protal \& splenic veins, \legendsquare{con12} pancreas, \legendsquare{con13} right adrenal gland, and \legendsquare{con14} left adrenal gland.}
 \vspace{-3mm}
 \label{fig:result}
\end{figure*}

\subsection{Datasets and implementation}
\subsubsection{Datasets.} 
We use Synapse and AMOS datasets to evaluate our approach. 
Please refer to the supplementary material for further details.



\subsubsection{Implementation details.} 
All experiments are implemented in PyTorch~2.6.0 (CUDA~11.8 build) with an EMA teacher--student framework, trained on a single NVIDIA RTX~4090~D GPU. 
We use SGD with momentum~0.9 and weight decay $1\times 10^{-4}$. The learning-rate follows a polynomial schedule with $\text{base\_lr}=0.01$ and 0.9 pow. Each mini-batch has $4$ volumes with $2$ labeled and $2$ unlabeled. 
At each iteration we randomly crop a $96\times 96\times 96$ subvolume. 
We sample parameters for \emph{weak} and \emph{strong} 3D augmentations and apply them consistently: 
labeled images and their masks are augmented in both weak/strong branches; unlabeled images are augmented to produce weak/strong inputs.
For teacher predictions on unlabeled data, we add a small Gaussian perturbation to inputs (clamped noise) before forwarding the EMA model. 
Labeled samples are supervised by ground-truth masks. 
Unlabeled samples are supervised by EMA pseudo labels; the consistency weight follows a sigmoid ramp-up to $\lambda_u=1.0$ over the first $17{,}000$ iterations, and the EMA decay is $\omega_{\text{ema}}=0.99$. 
For the Slice-Block Shuffle path, supervised loss averages cross-entropy, GA-Dice, and Dice on the recovered head. 
Unsupervised consistency uses Dice between student predictions and the EMA teacher’s pseudo labels. 
For Confidence-Guided Displacement, we apply the hybrid CE\,+\,Dice objective defined in Sec.~\ref{subsec:conf_disp} to the displaced pairs. 
The confidence-guided displacement term is further weighted by $\lambda_{\text{disp}}$ and the current consistency weight. 
We set the cube/slice-block size $p=2$, Top-$K=2$, and $\lambda_{\text{disp}}=0.25$. The same randomly chosen axis is reused across weak/strong streams within an iteration.

\paragraph{Inference and evaluation.}
We perform sliding-window inference with stride $32\times 32\times 16$.
We report Dice~(\%) and Average Surface Distance (voxel) on the test set; for Synapse we additionally average results over three different seeds.

\subsection{Comparison with State-of-the-art Methods}
We compare JanusNet with semi‑supervised segmentation baselines, including general methods (UA‑MT~\cite{yu2019ua-mt}, URPC~\cite{luo2021urpc}, CPS~\cite{chen2021cps}, SS‑Net~\cite{wu2022ss-net}, DST~\cite{chen2022dst}, DePL~\cite{wang2022depl}, MagicNet~\cite{basak}) and approaches explicitly addressing class imbalance (Adsh~\cite{guo2022adsh}, CReST~\cite{wei2021crest}, SimiS~\cite{chen2022simis}, CLD~\cite{lin2022cld}, GenericSSL~\cite{wang2023genericssl}, SKCDF\cite{zhang2025semantic}, GA-MagicNet\cite{qi2024gradient}).

As presented in Tab.~\ref{with sota on synapse}, general semi‑supervised methods are unstable on small organs (some classes even approach zero Dice), and class‑imbalance designs help in part but still fail on structures such as \emph{esophagus} and \emph{adrenal glands}.
By contrast, JanusNet achieves the best Avg. Dice of 72.67\%, outperforming the strong baseline GA‑MagicNet 68.43\% by \textbf{+4.24}. 
Per‑organ, JanusNet yields large gains on challenging or small structures, such as stomach (+12.3\%), pancreas (+11.6\%), spleen (+5.4\%).

\begin{table*}[!t]
  \centering
   \footnotesize
   \setlength{\tabcolsep}{1.5mm}
   \begin{tabular}{@{}cccc|cc|cccccccccccccc@{}}
    \toprule

   & \multirow{2}{*}{Aug.} & \multirow{2}{*}{SBS} & \multirow{2}{*}{CGD} & Avg. & Avg. & \multicolumn{13}{c}{Dice of Each Class}  \\
   & &  & & Dice & ASD & Sp & RK & LK & Ga & Es & Li & St & Ao & IVC & PSV & Pa & RAG & LAG\\
   \midrule
   &            &            &            & 64.34 ± 0.58 & 3.34 ± 1.34 & 85.8 & 90.1 & 89.1 & 22.8 & 39.8 & 90.5 & 63.1 & 80.0 & 79.5 & 61.4 & 41.5 & 34.8 & 58.2 \\
   & \checkmark &            &            & 69.67 ± 0.69 & 3.54 ± 1.72 & 84.9 & 93.1 & 91.5 & 29.2 & 49.4 & 92.5 & 67.8 & 80.8 & 79.5 & 70.0 & 53.9 & 57.2 & 56.0 \\
   & \checkmark & \checkmark &            & 72.51 ± 0.54 & 3.96 ± 1.29 & 88.5 & 89.2 & 89.6 & 41.5 & 57.4 & 92.5 & 72.9 & 79.4 & 83.6 & 70.9 & 60.6 & 54.4 & 62.1 \\ 
   & \checkmark &            & \checkmark & 71.82 ± 1.54 & 3.92 ± 1.62 & 90.6 & 89.6 & 91.1 & 39.4 & 54.7 & 94.7 & 73.2 & 82.0 & 82.2 & 70.7 & 62.6 & 44.8 & 58.1 \\
   & \checkmark & \checkmark & \checkmark & 72.67 ± 1.18 & 3.82 ± 0.47 & 87.9 & 90.2 & 90.1 & 40.7 & 55.0 & 93.3 & 75.0 & 79.2 & 83.3 & 71.4 & 62.5 & 55.7 & 60.5 \\ 
   
   \bottomrule
  \end{tabular}
  \vspace{-2mm}
  \caption{Ablation study on the 20\% labeled Synapse dataset to evaluate the effectiveness of each component. Aug.: weak \& strong augmentation streams. SBS: slice-block shuffle step. CGD: confidence-guided displacement step. }
  \label{ablation main}
  \vspace{-2mm}
\end{table*}

\begin{table}[t]
  \centering
  \footnotesize
   \begin{tabular}{@{}cl|ccc@{}}
    \toprule
     \multicolumn{2}{c|}{$p$} & Avg. Dice & Avg. ASD &  \\
     \midrule
     & 2 & 72.48 ± 1.52 & 3.92 ± 2.58 \\
     & 4 & 72.50 ± 1.01 & 4.21 ± 1.60 \\
     & 8 & 72.60 ± 0.14 & \textbf{3.82 ± 0.12} \\
     & 16 & \textbf{72.67 ± 1.18} & \textbf{3.82 ± 0.47} \\
     & 32 & 72.14 ± 1.21 & 3.47 ± 1.57 \\
    \bottomrule
  \end{tabular}
  \vspace{-2mm}
  \caption{Ablation study on the effect of slice-block thickness $p$ on the 20\% labeled Synapse dataset.  }
  \label{ablation p}
  \vspace{-2mm}
\end{table}

\begin{table}[t]
  \centering
  \footnotesize
   \begin{tabular}{@{}cl|ccc@{}}
    \toprule
     \multicolumn{2}{c|}{$\lambda_{\text{disp}}$} & Avg. Dice & Avg. ASD &  \\
     \midrule
     & 0.00 & 72.51 ± 0.54 & 3.96 ± 1.29 \\
     & 0.25 & \textbf{72.67 ± 1.18} & \textbf{3.82 ± 0.47} \\
     & 0.50 & 72.50 ± 1.01 & 3.98 ± 1.60 \\
     & 0.75 & 72.16 ± 0.14 & 4.28 ± 0.12 \\
     & 1.00 & 71.40 ± 1.21 & 4.39 ± 1.57 \\
    \bottomrule
  \end{tabular}
  \vspace{-2mm}
  \caption{Ablation study on the displacement loss weight $\lambda_{\text{disp}}$ on the 20\% labeled Synapse dataset.  }
  \label{ablation lambda disp}
  \vspace{-3mm}
\end{table}

\begin{table}[t]
  \centering
  \footnotesize
   \begin{tabular}{@{}cl|ccc@{}}
    \toprule
     \multicolumn{2}{c|}{Top-$K$} & Avg. Dice & Avg. ASD &  \\
     \midrule
     & 1 & 72.10 ± 0.20 & 3.87 ± 1.14 \\
     & 2 & \textbf{72.67 ± 1.18} & \textbf{3.82 ± 0.47} \\
     & 3 & 72.12 ± 0.31 & 3.86 ± 0.52 \\
     & 4 & 71.56 ± 0.55 & 4.14 ± 1.53 \\
     & 5 & 71.80 ± 1.02 & 4.31 ± 1.91 \\
    \bottomrule
  \end{tabular}
  \vspace{-2mm}
  \caption{Ablation study on the Top-$K$ selection strategy on the 20\% labeled Synapse dataset. }
  \label{ablation topk}
  \vspace{-4mm}
\end{table}

Table~\ref{with sota on amos} summarizes the comparison on the AMOS dataset (5\% labels). With only 5\% annotations, JanusNet attains 63.99\% Avg.~Dice and 4.45 Avg.~ASD—both the best—and shows clear advantages over imbalance‑aware competitors. A key strength of JanusNet is segmenting elongated, boundary‑ambiguous, or context‑dependent organs. Compared with GA‑MagicNet it yields notable per‑class gains: Esophagus (+9.2\%), Stomach (+7.4\%), Duodenum (+6.3\%), IVC (+3.5\%), Pancreas (+2.5\%); it also improves large organs such as Spleen (+3.0\%) and Liver (+0.5\%). 

We further visualize the segmentation results of the various methods on the Synapse dataset, as illustrated in Fig.\ref{fig:result}. We can observe that other methods are more likely to over-segment or under-segment, and the segmentation target is more likely to have hollows. Our proposed JanusNet still shows better performance.

\subsection{Ablation Analysis}
\paragraph{Effectiveness of each step.}
We evaluate the contribution of each component, as presented in Tab.~\ref{ablation main}. The first row uses a Mean‑Teacher (MT) model with naive pseudo‑label supervision as the baseline. After introducing \emph{weak\&strong} augmentations, the Avg.~Dice improves by \(+5.33\%\). On top of this setting, adding \emph{Slice‑Block Shuffle} (SBS) alone and \emph{Confidence‑Guided Displacement} (CGD) alone brings additional gains of \(+2.84\%\) and \(+2.15\%\), respectively. Combining SBS and CGD with the \emph{weak\&strong} baseline yields the best result of \(72.67\%\) DSC, surpassing the baseline by \(+8.33\%\). These improvements suggest that the two steps complement each other at different granularity levels: SBS injects anatomical priors at the slice‑block level to learn organs’ relative positions, while CGD emphasizes hard regions at the patch level to improve discriminability.

\paragraph{Effect of Slice–Block thickness $p$.}
As presented in Tab.~\ref{ablation p}, increasing the block thickness from $p{=}2$ to $p{=}16$ steadily improves Avg.~Dice from 72.48 to 72.67, while Avg.~ASD decreases from 3.92 to 3.82, indicating the most accurate and stable setting. 
We attribute this to a balance between structure preservation and augmentation diversity: very small blocks (2/4) inject stronger recomposition but also higher cross‑voxel mismatch noise, slightly destabilizing boundaries; overly large blocks (32) over‑preserve local anatomy, weakening cross‑case perturbation and regularization, which reduces Avg.~Dice to 72.14 despite occasionally smoother boundaries (lowest ASD 3.47 with larger variance).
Considering both accuracy and stability, we adopt $p{=}16$ by default.

\paragraph{Effect of displacement loss weight $\lambda_{\text{disp}}$.}
Tab.~\ref{ablation lambda disp} shows that introducing a moderate displacement weight is beneficial: from $\lambda_{\text{disp}}{=}0$ to $0.25$, Avg.~Dice improves from 72.51 to 72.67, and ASD drops to 3.82.
Further increasing the weight (0.50/0.75/1.00) causes monotonic degradation (Dice from 72.50 to 71.40, ASD from 3.98 to 4.39), suggesting that over‑emphasizing displacement perturbs anatomical continuity and weakens teacher‑driven stabilization.
We therefore use $\lambda_{\text{disp}}{=}0.25$ as the default trade‑off to amplify hard‑region signals while avoiding over‑perturbation.

\paragraph{Effect of Top-$K$ for hard‑example selection.}
As reported in Tab.~\ref{ablation topk}, increasing $K$ from $1$ to $2$ boosts performance from 72.10 to 72.67 (ASD down to 3.82), after which larger $K$ (3/4/5) yields a downward trend (Dice down to 71.56, ASD up to 4.31).
This indicates that injecting a \emph{small yet precise} subset of hard instances best supports stable optimization: $K{=}1$ under‑covers difficult regions, whereas $K{\ge}3$ over‑rearranges blocks within a slice per iteration, undermining anatomical priors and amplifying gradient fluctuations.
We adopt \textbf{Top-$K{=}2$} as the default configuration.

\section*{Conclusion}

In this paper, we propose JanusNet, which provides more efficient data augmentation for semi-supervised 3D multi-organ segmentation. JanusNet adopts a teacher-student framework and includes two stage-wise, layer-aware augmentations built on the slice-block shuffle. The first stage enforces global layerwise alignment, and the second stage performs local in-layer refinement. The two stages act progressively and cooperatively, striking a balance between global structure and difficult local details. 
Extensive experiments on the Synapse and AMOS datasets demonstrate that JanusNet significantly surpasses state-of-the-art methods. 

\bibliography{aaai2026}

\end{document}